\documentclass[10pt,twocolumn]{article}

\usepackage[utf8]{inputenc}
\usepackage[T1]{fontenc}
\usepackage{lmodern}
\usepackage[letterpaper,margin=0.75in,columnsep=0.25in]{geometry}
\usepackage[numbers,sort&compress]{natbib}
\usepackage{amsmath,amssymb,array}
\usepackage{bm}
\usepackage{graphicx}
\usepackage{xcolor}
\usepackage{booktabs}
\usepackage{multirow}
\usepackage{listings}
\usepackage[hyphens]{url}
\usepackage{microtype}
\usepackage[font=small,labelfont=bf]{caption}
\usepackage{authblk}
\usepackage[colorlinks=true,linkcolor=blue!60!black,citecolor=blue!60!black,urlcolor=blue!60!black]{hyperref}
\newcommand{\Fnorm}[1]{\left\lVert #1 \right\rVert_F^2}
\newcommand{\lr}[1]{\left( #1 \right)}

\newcommand{\diag}[1]{\operatorname{diag}\!\left( #1 \right)}

\newcommand{\norm}[1]{\left\lVert #1 \right\rVert}
\newcommand{\inner}[2]{\langle #1, #2 \rangle}
\newcommand{\proj}[1]{P_{#1}}

\title{FermiGrad-SVD: An LLM Compression Pipeline with Global Rank Selection and Gauge Fixing}

\author[1]{Roman Rausch\thanks{Corresponding author: \href{mailto:roman.rausch@multiversecomputing.com}{\protect\nolinkurl{roman.rausch@multiversecomputing.com}}}}
\author[1]{David Jansen}
\author[2]{Sukhbinder Singh}
\author[1,3,4]{Rom\'an Or\'us}
\affil[1]{Multiverse Computing, Paseo de Miram\'on 170, 20014 Donostia, Spain}
\affil[2]{Multiverse Computing, 192 Spadina Avenue, Toronto, ON M5T 2C2, Canada}
\affil[3]{Donostia International Physics Center, San Sebasti\'an, Spain}
\affil[4]{Ikerbasque Foundation for Science, Bilbao, Spain}
\date{}

\begin{document}

\maketitle

\begin{abstract}
Low-rank factorization of weight matrices is a promising approach to compress large language models (LLMs), but its practical effectiveness depends on allocating ranks across layers and efficiently representing the resulting factors. We introduce three complementary methods for improving SVD-based LLM compression. First, \textbf{FermiGrad} formulates global layer-wise rank allocation as a differentiable optimization problem by replacing discrete singular-value truncation with continuous Fermi gates. Because FermiGrad optimizes only rank parameters over precomputed low-rank factors, it avoids repeated activation decompositions and backpropagation through SVD operations. Second, the \textbf{Targeted Compressor} extends FermiGrad with a task-specific objective that directs limited model capacity toward a selected domain.  At 60\% parameter density for Llama-3.1-8B-Instruct, SciQ-targeted compression achieves 60\% MMLU accuracy, exceeding the reported quantized Dobi-SVD result by 10 percentage points, while deliberately trading performance on unrelated tasks for domain specialization. Third, \textbf{PivGa} provides a lossless secondary compression of low-rank factors by exploiting the gauge freedom in their parametrization. The proposed framework is modular, computationally efficient, and applicable at scale, as demonstrated by compressing Llama-3.3-70B-Instruct to 50\% parameter density.
\end{abstract}

\section{Introduction}

Large Language Models (LLMs) demonstrate remarkable linguistic capabilities, allowing them to be widely deployed as translators, tutors, and personal assistants. Achieving such performance, however, requires models with billions to trillions of parameters, demanding substantial computational resources. In particular, their memory footprint makes it difficult to run them on smaller or edge devices. These constraints have motivated the development of a range of LLM compression techniques. The range of methods includes: quantization (reduction in precision)~\cite{Dettmers2022_quant,frantar2022gptq,lin2023awq}, block removal (elimination of whole transformer blocks)~\cite{Gromov2024_Unreasonable,Sreenivas2024_Minitron,jansen2026} and pruning (reduction of matrix dimensions)~\cite{sun2023wanda,muralidharan_24,guo2025slimllm,Gili_25,Sreenivas2024_Minitron}. Finally, low-rank techniques decompose the weights as a contraction of smaller tensors~\cite{Novikov2015_TensorizingNN}. In this context, the simple Singular Value Decomposition (SVD) has particular appeal, since it is known to produce the optimal low-rank approximation of the original matrix, and has been a focus of research for LLM compression~\cite{Yuan2023_ASVD,FWSVD,Wang2024_SVD-LLM,Zhao2025_Pifa,Wang2025_DobiSVD,ParkinaRakhuba2025_COALA}.

\begin{figure}[!ht]
    \centering
    \includegraphics[width=0.7\columnwidth]{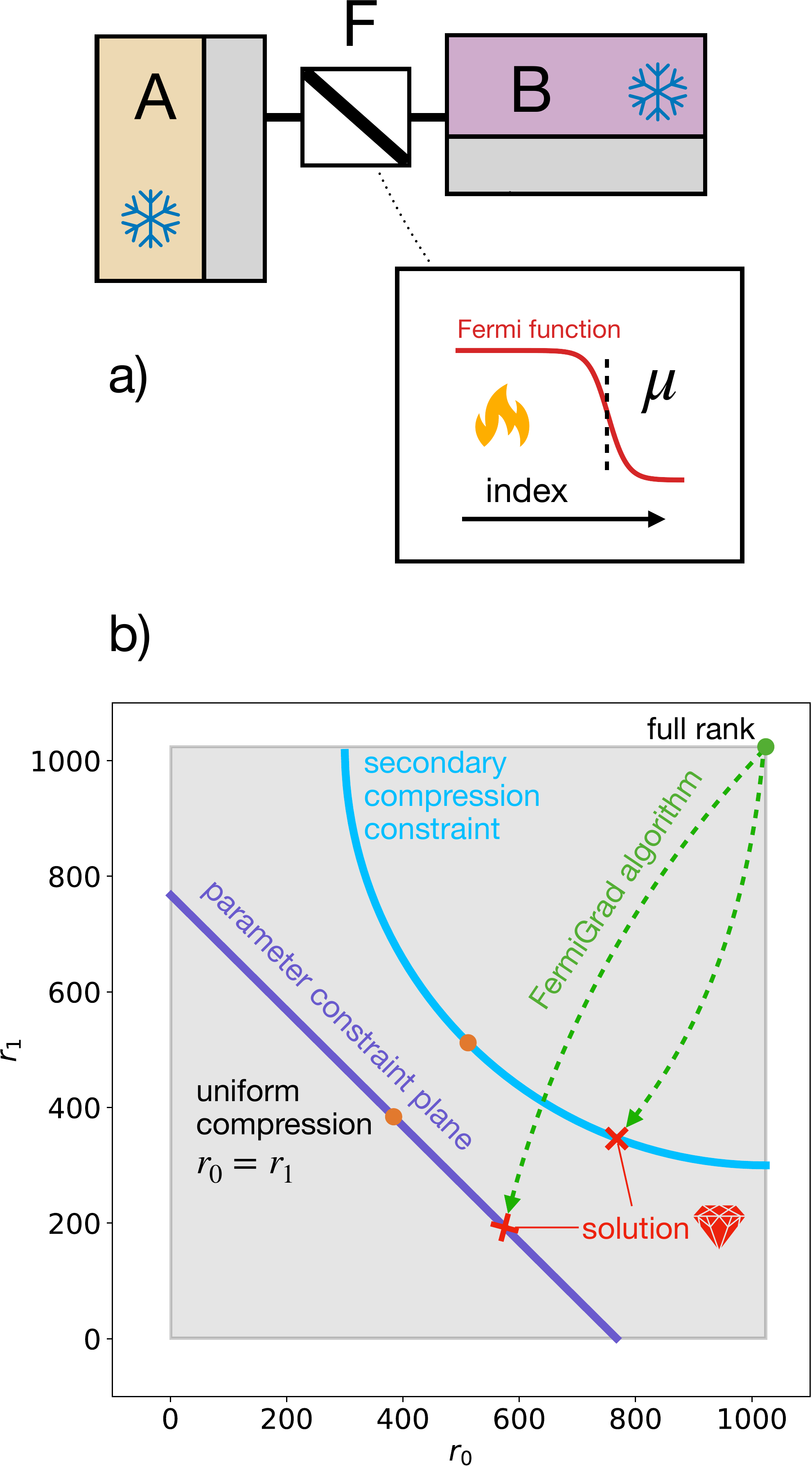}
    \caption{
    a) Illustration of FermiGrad for a single layer: $F$ is the active Fermi tensor that softens the SVD truncation while $A$ and $B$ are frozen, and $\mu$ provides a slider to select the singular values.
    b) Optimization space for two 1024$\times$1024 layers (2M parameters) and 1.57M target parameters. The grey area shows the box constraint for the ranks: $1\leq r_l\leq1024$, $l=0,1$.
    The red crosses denote the hypothesized optimal solution. The blue (light blue) lines show the parameter constraints for regular (secondary) compression. For the regular constraint, the FermiGrad algorithm starts with full rank and moves along some trajectory to the optimized solution as the penalty term is increased (see text).
    }
    \label{fig:fermigrad}
\end{figure}

\subsection{\label{sec:related}Preceding Work}

It was soon realized that naive SVD performs poorly when simply applied to LLM model weights because small singular values may actually contribute significantly to the downstream model output~\cite{FWSVD}. To account for this, additional importance weighting from data is crucial. Another issue is the combinatorial problem of selecting the optimal SVD ranks for each layer. A third issue is dealing with the factorization overhead of the SVD. A difficulty in comparing methods is that published works tend to address several of these problems in a single package, even though one could mix-and-match different weighting, rank-finding and overhead-reduction techniques.

The first work that recognized the importance of weighting was FWSVD using Fisher importance~\cite{FWSVD}. This was done only in an approximated form, scaling the weights by a diagonal matrix.
ASVD also used a diagonal matrix with values that derive from activation statistics~\cite{Yuan2023_ASVD} and proposed a sensitivity-based rank searching (STRS), whereby a global performance drop is measured by compressing individual layers to different compression ratios.
SVD-LLM~\cite{Wang2024_SVD-LLM} directly improved upon the ASVD approach by generalizing to a nondiagonal calibration matrix, and putting the theory on a firmer footing. This calibration technique could outperform previous methods even without special rank selection.
Our main contribution, the \textbf{FermiGrad} algorithm, lies in this line of work and extends SVD-LLM by adaptively choosing the ranks of each SVD layer by optimizing a global objective. With a more specialized loss function, FermiGrad becomes the \textbf{Targeted Compressor} and can effectively preserve specific knowledge in the smaller model.

Recently, IO-SVD~\cite{abbasi2026iosvd} appeared as a different improvement over SVD-LLM. It uses a double-sided calibration (over inputs and outputs), bundled together with a rank-finding greedy algorithm.  We note that the double-sided calibration is fully compatible with our FermiGrad algorithm, which could take such a model as input.

A different line of development was pursued by Dobi-SVD~\cite{Wang2025_DobiSVD}, which performs SVD on the activations, and then finds a decomposition of the weights based on IPCA (incremental principal components analysis). To find the optimal rank for each layer, Dobi-SVD introduces a differentiable soft constraint, which is similar to our FermiGrad algorithm. However, since it is applied to the activations, it needs to perform SVD with backpropagation in float32, which makes it quite costly (while also dealing with instabilities). We will show that FermiGrad is much more data-efficient, and the Targeted Compressor can outperform Dobi-SVD on the targeted MMLU benchmark.

The ``factorization overhead'' of SVD is the problem that large ranks increase rather than decrease the number of parameters (cf. Sec.~\ref{sec:pivga}). Dobi-SVD and IO-SVD deal with this by quantizing the decomposed weights (termed ``remapping''); the former statically, the latter in an adaptive fashion. However, this means that they use two orthogonal compression techniques on top of each other.
On the other hand, the Pivoting Factorization (PiFa)~\cite{Zhao2025_Pifa} method introduces a secondary compression that removes the factorization overhead at the price of slower inference. We improve on this method by introducing an approach we call \textbf{PivGa} (for ``pivoted gauge''), which is based on interpolative decomposition and ameliorates the speed loss. This is the secondary contribution of our paper.

\section{\label{sec:svd}Data-aware SVD compression}

The SVD decomposes a rectangular $m\times n$ matrix (which we assume to have full rank) as
\begin{equation}
W=U\Sigma V^T,
\label{eq:SVD}
\end{equation}
where $\Sigma$ is the diagonal matrix of descending singular values. $U$ and $V$ have orthonormal columns, $U^T U = I_q$, $V^T V = I_q$ ($q=\min\lr{m,n}$), and $U U^T$ ($V V^T$) is the projector on the column (row) space of $W$.
If $\Sigma_r$ denotes the top-$r$ singular values, then
$W = U_r \Sigma_r V_r^T + U_\perp \Sigma_\perp V_\perp^T = W_r + W_{\perp}$. 
SVD \textit{compression} consists in only keeping $W_r$ and neglecting $W_{\perp}$. By the Eckart-Young Theorem, $W_r = U_r \Sigma_r V_r^T$ is the optimal rank-$r$ approximation of $W$ in terms of the Frobenius norm, namely, $W_r \in\underset{\operatorname{rank}(W') \le r}{\arg\min}\;\Fnorm{W-W'}$.
A useful property is that $W_r$ can also be obtained by a left(right)-projection $W_r=U_rU_r^TW$ ($W_r=WV_rV_r^T$). 

In the context of LLMs, a naive SVD compression of the weights is not aligned with the task of optimizing the downstream model performance~\cite{FWSVD}.
A better aligned objective is instead to minimize $\mathcal{L} = \sum_b \Fnorm{WX_b - ABX_b}$~\cite{Wang2024_SVD-LLM}, where $X_b$ is a batch of a calibration data. Expanding the square in this loss function naturally leads to the introduction of the calibration matrix $C = \sum_b X_b X_b^T$. We can decompose $C$ into a product of square matrices $C=SS^T$ (provided positive definiteness), and then re-write 
\begin{equation}
\mathcal{L} = \Fnorm{WS-ABS},
\end{equation} 
so that $W$  is weighted by $S$. This objective now supplants Eq.~\eqref{eq:SVD} for LLMs.
By the Eckart-Young Theorem, the minimum of $\mathcal{L}$ subject to rank $r$ is obtained by the truncated rank-$r$ SVD of $WS$, i.e. $svd(WS)_r = U_r\Sigma_r V_r^T = ABS$. We can thus set $A=U_r$, $B=U_r^T W$, whereby using the left-projected formulation avoids an explicit inverse computation of $S$~\cite{ParkinaRakhuba2025_COALA}. To obtain $S$ from $C$, we use the Cholesky decomposition $C=LL^T \Rightarrow S=L$ as in the original proposal, for which we find no speed or stability issues in our calibration datasets, unlike reported by other authors~\cite{ParkinaRakhuba2025_COALA}.

\section{Secondary compression: Pivoted Gauge Fixing (PivGa)} \label{sec:pivga}

The number of parameters after the SVD truncation is $r\lr{n+m}$, which is only smaller than the original size $mn$ if $r$ is smaller than $r_b = mn/\lr{m+n}$ (``breakeven rank''). However, one can further reduce the parameter count without additional approximations, by exploiting the ``gauge freedom'' implicit in a low-rank decomposition~\cite{Koike-Akino2025_LatentLLM}. Namely, we can insert any invertible $r\times r$ matrix between $A$ and $B$, $AB\to AG^{-1}G B$. In particular, we can try choosing $G$ to invert the first $r\times r$ block of $B$, denoted $B_0$, so that
$
AB = \lr{A B_0} \begin{bmatrix}I_{r} & B_0^{-1}B_1\end{bmatrix}
= A' B'
$.
Since the ($r \times r$) identity matrix $I_{r}$ need not be stored explicitly, the parameter count reduces to $r(n+m)-r^2\leq mn$. However, the block $B_0$ is typically ill-conditioned, and using an explicit inverse destroys the model.

It has been noted that this issue can be circumvented using pivoting~\cite{Zhao2025_Pifa,Koike-Akino2025_LatentLLM}. Here, we present an approach which is similar to the \textit{Pivoting Factorization} (PiFa)~\cite{Zhao2025_Pifa}, but is based on the \textit{Interpolative Decomposition} (ID). Let us permute the columns of $W$ so that the first $r$ columns are linearly independent (the ``skeleton columns''), $W\Pi = \begin{bmatrix}C & W_{\text{rest}}\end{bmatrix}$. $\Pi$ is a permutation matrix, which can be determined using LU decomposition as explained in~\cite{Zhao2025_Pifa}. Performing the QR decomposition and selecting a block $R_{11}\in \mathbb{R}^{r\times r}$ of $R$, we get
$W\Pi = \begin{bmatrix}Q_1 & Q_2\end{bmatrix} \begin{bmatrix}R_{11} & R_{12} \\ 0 & R_{22}\end{bmatrix}$.

If $W$ has exact rank $r$, we have $R_{22}=0$ and the skeleton columns become
$C = Q_1 R_{11} \in \mathbb{R}^{m\times r}$. The remaining columns are expressed
in the skeleton-column basis through the coefficient matrix
$\tilde D = \begin{bmatrix}I_{r} & D\end{bmatrix}\in\mathbb{R}^{r\times n}$,
obtained from the triangular system
$R_{11}\tilde D = \begin{bmatrix}R_{11} & R_{12}\end{bmatrix}$
(see Appendix~\ref{app:pivga} for the derivation). Its first block returns
$I_r$, while its second block is the stored quantity
$D = R_{11}^{-1}R_{12}\in\mathbb{R}^{r\times(n-r)}$. The weight matrix is then
expressed as $W = C\,\tilde D\,\Pi^{-1} = C\begin{bmatrix}I_{r} & D\end{bmatrix}\Pi^{-1}$.

\begin{figure}[t]
    \centering
    \includegraphics[width=0.7\columnwidth]{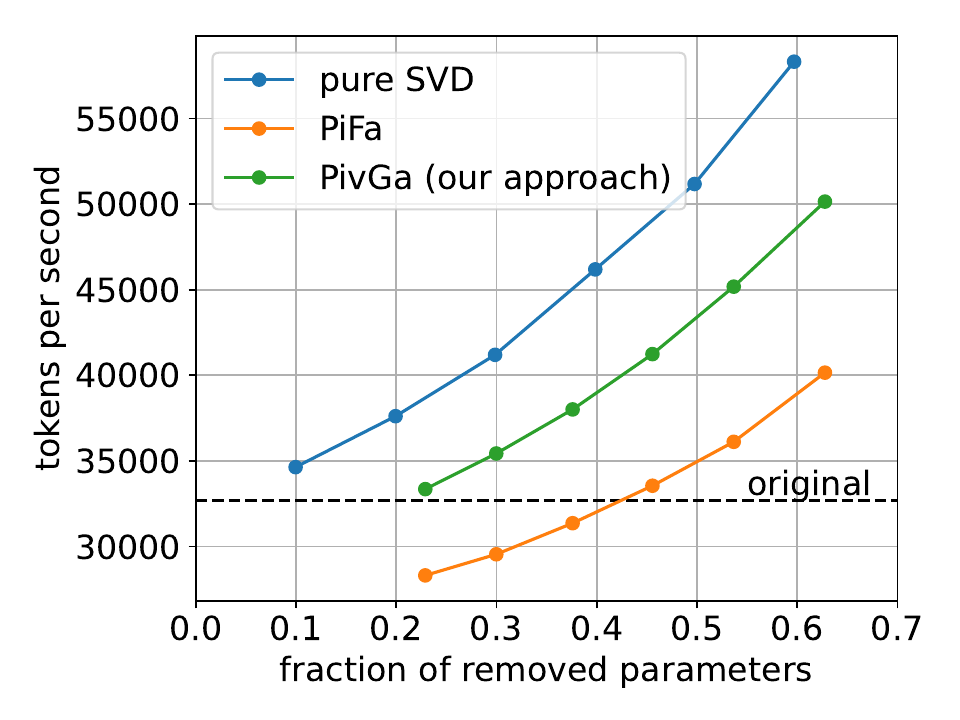}
    \caption{Inference speed of the PivGa approach compared to PiFa~\cite{Zhao2025_Pifa} and the pure SVD model (without PiFa/PivGa) for random tokens with sequence length 256, batch size 32 on an H200 GPU using Llama-3.1-8B-Instruct in bfloat16.}
    \label{fig:pivga}
\end{figure}

\begin{figure}[t]
    \centering
    \includegraphics[width=\columnwidth]{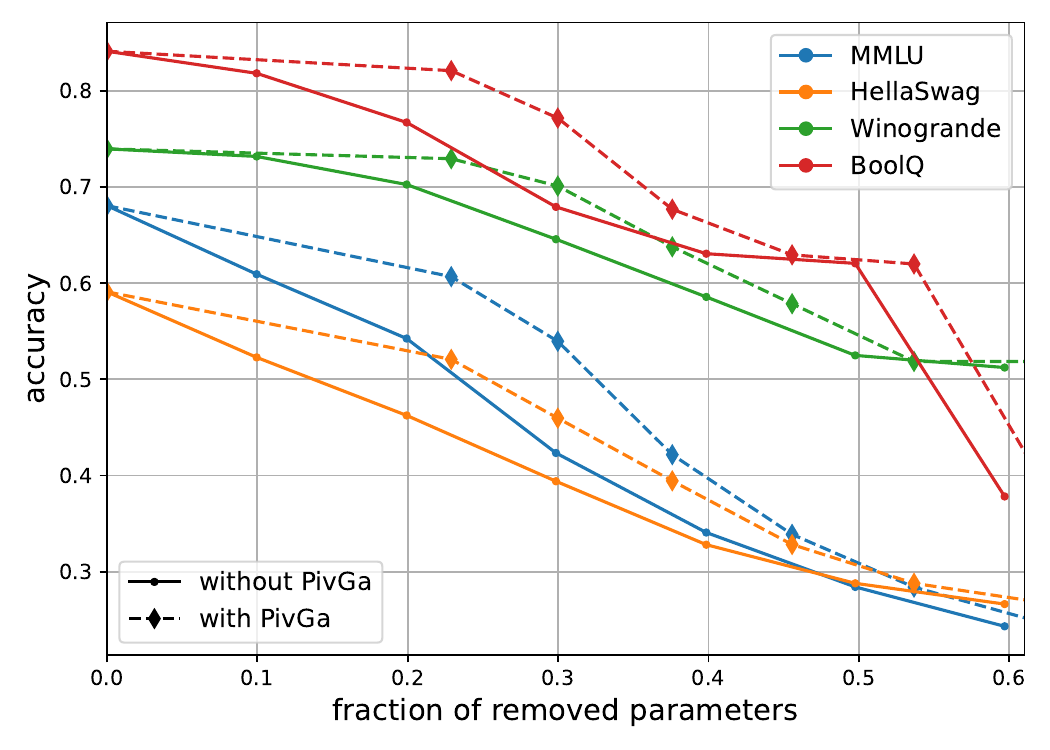}
    \caption{
    Change in the accuracy vs. parameter-count curves when secondary compression (PivGa) is applied. This secondary compression is lossless, and so the benchmark curves essentially shift sideways, with a larger shift at lower compression rates.
    }
    \label{fig:pivga_accuracy}
\end{figure}

Our approach can be implemented in PyTorch in a faster way than PiFa, see Fig.~\ref{fig:pivga}. The forward pass is implemented as follows:
\begin{lstlisting}
import torch.nn.functional as F
def forward(self, x): 
  idx = self.Pi.expand(*x.shape[:-1], -1) # apply permutation
  x_perm = x.gather(-1, idx) 
  x1 = x_perm[..., :self.rank] # identity block
  x2 = x_perm[..., self.rank:] # D block
  return F.linear(x1 + F.linear(x2, self.D), self.C, self.bias)
\end{lstlisting}
Storing the permutation metadata has negligible effect on the total memory, but performing it in the forward pass makes the result slower than the pure SVD model. 
Essentially, the secondary compression is a way to trade speed for storage, and the utility of the method depends on the use case.
Across the measured densities, however, PivGa remains faster than the original (unlike PiFa in our experiments, cf. Fig.~\ref{fig:pivga}). 
Accuracy retention is demonstrated in Fig.~\ref{fig:pivga_accuracy}.

\section{Finding the ranks globally: FermiGrad}

The central limitation of data-aware SVD compression (Sec.~\ref{sec:svd}) is the selection of optimal layer-wise ranks $r_l$ for a given target model size. Rank selection is an optimization problem whose search space grows with the number of layers. Sensitivity approaches that profile each layer independently like STRS~\cite{Yuan2023_ASVD} circumvent this combinatorial difficulty, but have the shortcoming of neglecting inter-layer correlations, which is akin to a ``mean-field'' method in physics.

We propose a way to find a direct solution in the global layer space by using gradient descent. Directly truncating weights to specific ranks is a discrete, nondifferentiable operation. To overcome this, we introduce a soft truncation of the weights that is approximate, but differentiable. 
We replace the low-rank decomposition $W_l\to A_lB_l$ by $W_l\to A_lF^{(l)}B_l$, where $F^{(l)}=\diag{F_0^{(l)},F_1^{(l)},\ldots}$ is a diagonal Fermi gate (cf. Fig.~\ref{fig:fermigrad}). 
To approximate the hard truncation, $F^{\lr{l}}_j$ must be a smooth function that is close to 1 for $j<r$ and close to 0 for $j>r$. A function that fulfills this requirement is the Fermi function from condensed-matter physics\footnote{Also known as the Woods-Saxon potential in nuclear physics.}. We use the rescaled form
\begin{equation}
F_{j}^{\lr{l}} =
\left[
1+\exp\left(\frac{j-\mu_l}{N_lT}\right)
\right]^{-1},
\qquad
j=0,\ldots,N_l-1,
\end{equation}
where $T$ is the temperature that controls the width of the transition, $\mu_l$ is the ``chemical potential'' at layer $l$ that continuously slides the truncation position, and $N_l$ is the maximum rank. We train the approximated model with frozen weights and active $\mu_l$; ultimately making a hard truncation at the learned ranks $r_l$ by rounding $\mu_l$ to the nearest integer.
We set the temperature $T=0.01$, which we found empirically to be small enough to provide a good approximation while still large enough to avoid numerical instabilities.

The gradient-descent optimization has furthermore two constraints. First, a box constraint: 
The ranks (and hence the $\mu_l$) lie between a minimum rank $r_\text{min}$ (we set $r_{\text{min}}=64$) and $N_l=\text{min}\lr{n_l,m_l}$, where $m_l$ ($n_l$) is the number of rows (columns) of layer $l$. And second, the (softened) target parameter count is fixed at $N_{\text{param}} = \sum_l \mu_l \lr{n_l+m_l} + N_{\text{inc}} =: \vec{\mu}\cdot\vec{a}+b$, where $N_{\text{inc}}$ is the number of all parameters that we ignore for the compression (such as the normalization and embedding layers). This is a hyperplane constraint. If we apply secondary compression (see Sec.~\ref{sec:pivga}), the parameter count changes to a quadratic constraint: $N_{\text{param}} = \sum_l \left[ \mu_l \lr{n_l+m_l} -\mu_l^2\right] + N_{\text{inc}} = \vec{\mu}\cdot\vec{a}-\lVert\vec{\mu}\rVert_2^2+b$.

For the optimization we choose a knowledge distillation loss function
$\mathcal{L} = D_{KL} + \mathcal{L}_{\text{constr}}$ where $D_{KL}$ is the
Kullback-Leibler (KL) divergence between the uncompressed (teacher) model and the
compressed (student) model, $D_{KL}\lr{P_{\text{teach}}||P_{\text{stud}}}$, and
$\mathcal{L}_{\text{constr}}$ is a penalty term that enforces the parameter count
constraint,
\begin{equation}
\mathcal{L}_{\text{constr}}
= \frac{\rho}{2}
  \lr{\frac{N_{\text{param}}\lr{\vec{\mu}} - N_{\text{target}}}{N_{\text{scale}}}}^{2}.
\end{equation}
The violation of the parameter count constraint is regulated by the hyperparameter
$\rho$. The constraint is exactly fulfilled for $\rho\to\infty$ under standard regularity and convergence assumptions, but in practice sufficiently large values of $\rho$ provide a good solution. We increase $\rho$ as a
function of the iteration step $t$ using the following protocol:
$\rho(t)=\min\lr{\rho_0 \alpha^t, \rho_{\text{max}}}$. We set $\rho_0=1.0$,
$\alpha\in[1.01,1.05]$, $\rho_{\text{max}}=2000.0$ and use the Adam optimizer with a learning rate of $lr=50$\footnote{Note that this is small on the scale of the matrix sizes, which is $O\lr{10^3}$ to $O\lr{10^4}$.}. The parameter $N_{\text{scale}}$ is set to a fixed
fraction of the full-rank (initial) parameter count, $N_{\text{scale}}=N_{\text{init}}/8$,
where $N_{\text{init}}$ is the total parameter count of the uncompressed model. The starting point is the full-rank model (cf.\ Fig.~\ref{fig:fermigrad}). The penalty term drives the model towards the parameter count constraint. After each iteration we clamp $\vec{\mu}$ to ensure the box constraint.

Our primary test model is Llama-3.1-8B-Instruct. The calibration stage is the same as SVD-LLM, but we have the budget to use as many as 65536 sequences to eliminate any potential under-convergence, which takes $\sim$10h on a single H200 GPU with batch size 128. The calibrated model is decomposed with full rank and cached on disk. FermiGrad then uses 1024 sequences for each step, and we precompute the teacher logits. For Llama-3.1-8B-Instruct, each iteration step takes $\sim$1min with batch size 4. The maximum number of iterations is $t_{\text{max}}=300$, and the full training finishes at most in $\sim$5.5 hours for a sequence length of 512. An early exit is implemented if the parameter constraint is reached earlier, which happens for shallow compression.

Using the softened truncation on the calibrated full-rank model, knowledge distillation loss and constraint ramping is what differentiates our method from Dobi-SVD. Our method is more efficient: Dobi-SVD applies the truncation to the activations and has to compute SVD with backpropagation in float32. It reports 8 GPU-hours for Llama-2-7B, but does not specify the hardware configuration or sufficient timing details to enable a normalized comparison. In our attempted reproduction on 4$\times$ H200 GPUs for Llama-3.1-8B-Instruct, the observed iteration cost projected to 7 days under the published hyperparameters.

\begin{table*}[tb]
\centering
\small
\setlength{\tabcolsep}{4pt}
\caption{Benchmark accuracy after compression with uniform SVD-LLM slicing (Unif), sensitivity-based rank allocation STRS~\cite{Yuan2023_ASVD}, and FermiGrad (FG) for the Llama-3.1-8B-Instruct model, shown for the two representative calibration datasets Tulu and MMLU-train across target densities (fraction of parameters retained). For each density and benchmark, the single best result across the three methods and both datasets is shown in bold. GSM8K uses strict-match exact-match; the remaining accuracies use standard accuracy. WikiText perplexity and the full comparison across all four calibration datasets are given in Tab.~\ref{tab:fermigrad_vs_uniform_full} (Appendix~\ref{app:full}).}
\label{tab:fermigrad_vs_uniform}
\begin{tabular}{llcccccccccccc}
\toprule
\multirow{2}{*}{density} & \multirow{2}{*}{Dataset} & \multicolumn{3}{c}{MMLU} & \multicolumn{3}{c}{HellaSwag} & \multicolumn{3}{c}{WinoGrande} & \multicolumn{3}{c}{GSM8K} \\
\cmidrule(lr){3-5}\cmidrule(lr){6-8}\cmidrule(lr){9-11}\cmidrule(lr){12-14}
 &  & Unif & STRS & FG & Unif & STRS & FG & Unif & STRS & FG & Unif & STRS & FG \\
\midrule
$1.0$ & Original & \multicolumn{3}{c}{67.8} & \multicolumn{3}{c}{59.1} & \multicolumn{3}{c}{73.6} & \multicolumn{3}{c}{75.9} \\
\midrule
\multirow{2}{*}{0.9} & Tulu & 53.0 & 55.6 & 59.1 & 47.3 & 44.0 & 50.2 & 68.3 & 66.9 & 70.2 & 30.9 & 23.8 & \textbf{46.3} \\
 & MMLU-train & 54.3 & 55.5 & \textbf{60.9} & 48.7 & 46.6 & \textbf{52.3} & 71.1 & 70.8 & \textbf{73.2} & 3.6 & 0.0 & 18.3 \\
\midrule
\multirow{2}{*}{0.8} & Tulu & 42.3 & 45.6 & 51.9 & 37.8 & 36.1 & 43.7 & 62.2 & 64.2 & 65.9 & 11.3 & 2.9 & \textbf{18.3} \\
 & MMLU-train & 45.4 & 48.7 & \textbf{54.2} & 41.1 & 38.3 & \textbf{46.2} & 66.3 & 67.4 & \textbf{70.2} & 0.0 & 0.0 & 0.3 \\
\midrule
\multirow{2}{*}{0.7} & Tulu & 31.7 & 33.3 & 38.1 & 30.7 & 29.9 & 36.4 & 54.7 & 57.3 & 61.1 & 1.7 & 0.0 & \textbf{5.2} \\
 & MMLU-train & 36.0 & 39.3 & \textbf{42.3} & 34.0 & 31.5 & \textbf{39.4} & 61.3 & 62.1 & \textbf{64.6} & 0.0 & 0.0 & 0.0 \\
\midrule
\multirow{2}{*}{0.6} & Tulu & 27.1 & 26.5 & 27.8 & 27.3 & 26.8 & 30.7 & 51.9 & 51.4 & 55.1 & 0.5 & 0.0 & \textbf{1.3} \\
 & MMLU-train & 29.1 & 29.5 & \textbf{34.1} & 28.8 & 27.6 & \textbf{32.8} & 54.5 & 56.0 & \textbf{58.6} & 0.0 & 0.0 & 0.0 \\
\midrule
\multirow{2}{*}{0.5} & Tulu & 25.8 & 23.0 & 24.6 & 26.6 & 26.2 & 28.3 & 49.5 & 51.9 & 51.2 & 0.0 & 0.0 & 0.0 \\
 & MMLU-train & 25.3 & 24.8 & \textbf{28.4} & 26.5 & 26.2 & \textbf{28.8} & 52.2 & \textbf{53.4} & 52.5 & 0.0 & 0.0 & 0.0 \\
\bottomrule
\end{tabular}
\end{table*}

\subsection{\label{sec:results}Results}

FermiGrad adds adaptive rank selection to SVD-LLM, hence our main baselines are the uniform-rank compression~\cite{Wang2024_SVD-LLM} and STRS~\cite{Yuan2023_ASVD}. Comparing to IO-SVD is not possible because their rank-selection method is bundled up with other double calibration and quantization, no code is provided as of the writing of this manuscript, and the authors limit themselves to the outdated Llama-2-7b model. Comparison to Dobi-SVD is only possible to few datapoints for Llama-3.1-8B-Instruct.

Previous works also usually limit themselves to the wikitext2 dataset~\cite{merity2017wikitext} for calibration and rank search. We find, however, that the end result is sensitive towards dataset selection, and that wikitext2 is a small, sub-quality dataset~\cite{jansen2026}. It seems more sensible to use a high-signal instruction dataset that one would also apply in fine-tuning. We show results for Tulu~\cite{muennighoff2024olmoe} and the train split of the mmlu~\cite{hendrycks2021mmlu}, for which we found the largest variance. Additional results for OpenHermes-2.5~\cite{teknium2023openhermes} and GeneralThought-420K~\cite{generalthought2025} are found in Appendix~\ref{app:full}.

Previous works fixated on the perplexity as the main loss metric. Perplexity always increases steeply for deeply compressed models, indicating loss of coherent text generation, but also recovers very quickly after only a few steps of retraining as coherence is recovered (cf. Sec.~\ref{sec:scalability}). For this reason, it is a more sensible metric only when comparing retrained models. Instead, we focus on the following benchmarks: MMLU (broad knowledge, multiple-choice), HellaSwag~\cite{zellers2019hellaswag} and WinoGrande~\cite{sakaguchi2020winogrande} (situational reasoning, multiple-choice), GSM8K~\cite{cobbe2021gsm8k} (high school math, must respond with numerical value). In particular MMLU is quite hard to recover with retraining on a small budget~\cite{jansen2026} (cf. Sec.~\ref{sec:scalability}) and is a valuable metric to preserve. All benchmarks are evaluated with the LM Evaluation Harness~\cite{lm-eval2021}.
We choose the sequence lengths according to the dataset statistics: 512 for MMLU, 1024 for the others. We filter out sequences that are longer than these caps to avoid interrupted texts. For a fair comparison with STRS, we use exactly the same loss function, dataset and sequence length as for FermiGrad. Each layer is compressed to densities [0.9...0.4], which requires $6 N_{\text{layers}}$ forward passes.

The results are shown in Tab.~\ref{tab:fermigrad_vs_uniform}. The FermiGrad approach clearly surpasses both uniform compression and STRS in almost all cases, while the precise choice of dataset allows to further adjust the type of the retained knowledge. 
For instance, Tulu is markedly better at preserving math ability (GSM8K), whereas MMLU-train yields the higher HellaSwag, WinoGrande, and MMLU scores. Overall, we find that MMLU-train is the best all-round dataset, especially considering its smaller sequence length. For clarity we show only these two representative datasets here; the full comparison across all four calibration datasets (additionally OpenHermes-2.5 and GeneralThought-420K) is provided in Appendix~\ref{app:full}.

Finally, we remark on the effect of secondary compression. While it is possible to directly target a parameter count that takes the secondary compression into account, this requires care: the gradient of the quadratic constraint, $n_l+m_l-2\mu_l$, vanishes at full rank for square layers ($n_l=m_l$), so these layers would never leave the starting point, and the optimization should instead be started slightly below full rank, $\mu_l=N_l-\delta$, so that the gradient does not vanish. It is more convenient to just apply PivGa to the end result, which is shown in Fig.~\ref{fig:pivga_accuracy}. The model forward pass is essentially unchanged after the transformation except for rounding errors, and we obtain the same benchmarks at a reduced parameter count.

\section{Preserving specific knowledge: Targeted Compressor}

A major use case for compressed models are narrow-domain experts that can be more capable in a specific domain than larger models. To create such a small expert, we are thus interested in preserving as much of that specific domain knowledge as possible during the compression.

To achieve this, we first pick a domain-specific multiple-choice dataset. Each text ends with the correct answer
letter (e.g.\ \texttt{"Answer: B"}), and the choices $\texttt{A},\texttt{B},\texttt{C},\texttt{D}$
are each mapped to a single vocabulary token. If we write $z_i^{\alpha}$ for the logit at the token option
$\alpha\in\{\texttt{A},\texttt{B},\texttt{C},\texttt{D}\}$, then the model's probability for option
$\alpha$ is the softmax over just the four choices,
\begin{equation}
p_i(\alpha)=\frac{e^{\,z_i^{\alpha}}}{\sum_{{\alpha}'\in\{\texttt{A},\texttt{B},\texttt{C},\texttt{D}\}} e^{\,z_i^{{\alpha}'}}},
\end{equation}
and the loss is the mean negative log-probability of the correct option $\alpha^{\text{ref}}_i$:
\begin{equation}
\mathcal{L}_{\text{classif}}=-\frac{1}{B}\sum_{i=1}^{B}\log p_i(\alpha^{\text{ref}}_i),
\end{equation}
where $B$ is the batch size. Because the softmax purposefully normalizes over the answer letters only, the model is rewarded purely
for ranking the correct letter above the other options. In the Targeted Compressor, this term replaces $D_{KL}$ as the
task loss, so FermiGrad now optimizes
$\mathcal{L}=\mathcal{L}_{\text{classif}}+\mathcal{L}_{\text{constr}}$.

As the specific datasets we choose MedMCQA~\cite{pal2022medmcqa} (medical), SciQ~\cite{welbl2017sciq} (natural sciences), and we compare with MMLU~\cite{hendrycks2021mmlu} (broad knowledge, no specific domain). We now benchmark the model also on the medical benchmarks MedMCQA and MedQA~\cite{jin2020disease}.

The results are shown in Tab.~\ref{tab:targeted} and we compare to the all-round best results using MMLU and KL divergence from before. We find that targeting MedMCQA indeed greatly improves the medical benchmarks compared to this baseline. It also tends to improve MMLU via the medical subsets. Targeting MMLU with classification loss is mostly worse than using it with KL divergence. The greatest impact, however, is achieved by targeting SciQ as classification: This gives by far the highest MMLU scores (mostly through MMLU domains overlapping with the SciQ dataset, like \textit{college biology} and \textit{professional medicine}), and even outperforms Dobi-SVD (roughly 5, 10, and 6 percentage points for 0.8, 0.6, and 0.4 density, respectively). Note that this is  despite the fact that Dobi-SVD uses quantization and re-defines ``model density'' as ``storage ratio'' instead of ``parameter ratio''.
Moreover, since the SciQ dataset has only short questions, we are very data-efficient and set the sequence length merely to 128, so that FermiGrad finishes in about $\sim$45min, keeping all other hyperparameters constant. The tradeoff of the Targeted Compressor is of course that the resulting model is only highly performant on the targeted domain, and not on general reasoning benchmarks like HellaSwag or WinoGrande.

\begin{table*}[!tp]
  \centering
  \small
  \setlength{\tabcolsep}{6pt}
  \caption{Benchmark accuracies as a function of the model density for the Targeted Compressor (TC-sciq, TC-mmlu and TC-medmcqa) compression approach. Bold marks the best result for each benchmark within each density block (across all methods shown).
  The FG-mmlu data are the same as in Tab.~\ref{tab:fermigrad_vs_uniform}.
    }
  \begin{tabular}{llccccc}
    \toprule
    density & Method & MMLU & MedMCQA & MedQA & HellaSwag & WinoGrande \\
    \midrule
    $1.0$ & Original & $67.8$ & $59.4$ & $63.7$ & $59.1$ & $73.6$ \\
    \midrule
    \multirow{4}{*}{$0.9$}
      & FG-mmlu & $60.9$ & $50.2$ & $52.5$ & $\mathbf{52.3}$ & $\mathbf{73.2}$ \\
      & TC-sciq & $\mathbf{66.4}$ & $56.9$ & $\mathbf{60.6}$ & $39.1$ & $69.7$ \\
      & TC-medmcqa & $64.8$ & $\mathbf{57.5}$ & $59.2$ & $33.8$ & $66.0$ \\
      & TC-mmlu & $59.5$ & $44.4$ & $36.5$ & $40.7$ & $71.0$ \\
    \midrule
    \multirow{5}{*}{$0.8$}
      & Dobi-SVD (quant.) & $60.1$ & -- & -- & -- & -- \\
      \cmidrule(lr){2-7}
      & FG-mmlu & $54.2$ & $38.9$ & $43.9$ & $\mathbf{46.2}$ & $\mathbf{70.2}$ \\
      & TC-sciq & $\mathbf{65.5}$ & $\mathbf{57.5}$ & $\mathbf{60.4}$ & $31.9$ & $65.5$ \\
      & TC-medmcqa & $62.3$ & $55.7$ & $50.3$ & $29.7$ & $58.9$ \\
      & TC-mmlu & $54.0$ & $40.0$ & $40.2$ & $30.8$ & $64.6$ \\
    \midrule
    \multirow{4}{*}{$0.7$}
      & FG-mmlu & $42.3$ & $31.3$ & $32.8$ & $\mathbf{39.4}$ & $\mathbf{64.6}$ \\
      & TC-sciq & $\mathbf{64.6}$ & $\mathbf{56.6}$ & $\mathbf{59.2}$ & $26.6$ & $58.0$ \\
      & TC-medmcqa & $51.0$ & $44.2$ & $35.5$ & $26.3$ & $52.5$ \\
      & TC-mmlu & $41.0$ & $29.0$ & $35.1$ & $28.2$ & $60.9$ \\
    \midrule
    \multirow{5}{*}{$0.6$}
      & Dobi-SVD (quant.) & $50.1$ & -- & -- & $\mathbf{52.0}$ & $\mathbf{72.0}$ \\
      \cmidrule(lr){2-7}
      & FG-mmlu & $34.1$ & $23.7$ & $24.0$ & $32.8$ & $58.6$ \\
      & TC-sciq & $\mathbf{60.3}$ & $\mathbf{53.3}$ & $\mathbf{56.6}$ & $25.6$ & $51.4$ \\
      & TC-medmcqa & $30.4$ & $28.9$ & $28.0$ & $25.9$ & $48.3$ \\
      & TC-mmlu & $35.3$ & $22.3$ & $24.7$ & $26.6$ & $55.0$ \\
    \midrule
    \multirow{4}{*}{$0.5$}
      & FG-mmlu & $28.4$ & $27.0$ & $29.0$ & $\mathbf{28.8}$ & $\mathbf{52.5}$ \\
      & TC-sciq & $\mathbf{54.3}$ & $\mathbf{44.9}$ & $\mathbf{53.2}$ & $26.1$ & $50.8$ \\
      & TC-medmcqa & $23.0$ & $32.2$ & $27.9$ & $26.1$ & $50.5$ \\
      & TC-mmlu & $23.0$ & $32.2$ & $24.9$ & $25.7$ & $51.0$ \\
    \midrule
    \multirow{5}{*}{$0.4$}
      & Dobi-SVD (quant.) & $28.2$ & -- & -- & $\mathbf{41.0}$ & $\mathbf{66.0}$ \\
      \cmidrule(lr){2-7}
      & FG-mmlu & $24.3$ & $28.4$ & $27.7$ & $26.6$ & $51.2$ \\
      & TC-sciq & $\mathbf{34.5}$ & $\mathbf{33.4}$ & $\mathbf{41.6}$ & $25.9$ & $48.0$ \\
      & TC-medmcqa & $23.0$ & $32.1$ & $27.7$ & $25.7$ & $49.6$ \\
      & TC-mmlu & $23.0$ & $31.6$ & $27.7$ & $25.5$ & $50.7$ \\
    \bottomrule
  \end{tabular}
  \label{tab:targeted}
\end{table*}

\section{\label{sec:scalability}Scalability}

FermiGrad involves model training with backpropagation. Even though all weights except for the $\vec{\mu}$ factors are frozen, it still raises the question of whether the method is scalable compared to sensitivity measures that only need forward passes without gradients.

We address this showing how FermiGrad scales to a production run for Llama-3.3-70B-Instruct (70.55B parameters, 80 blocks). (1) First, we run the recently developed block-removal algorithm~\cite{jansen2026}, which selects a high-quality solution of 24 blocks to remove. After this, the model shrinks to about 50B parameters, which is still quite significant. 
(2) Then, we run a brief QLoRA (4bit) supervised fine-tuning (SFT) recovery with $r=64$, $\alpha=128$, batch size 128, sequence length 1024 ($\sim$22k iterations), learning rate $10^{-4}$ on a mixture of GeneralThought, Tulu, OpenHermes-2.5 and TextbookReasoning~\cite{fan2025megascience}.
(3) We then calibrate the model using 1024 MMLU sequences, which takes $\sim$12h, and run FermiGrad only on the MLP layers with the target compression ratio of 50\%, which takes $\sim$22h. 
(4) Finally, we rebalance the low-rank factors (Appendix~\ref{app:balance}), freeze all weights except for the compressed MLP layers and perform a regular SFT recovery on the same dataset for 1 epoch (22k iterations, batch size 128, learning rate $5\cdot 10^{-5}$, sequence length 1024).
All these operations were performed on 4$\times$ H200 GPUs, and all trainings were performed using NeMo-AutoModel~\cite{nemo_automodel}. We note that this is a cheap budget for a model of that size.
The benchmarks after each stage are shown in Tab.~\ref{tab:llama70b-benchmarks}.

\begin{table*}[!t]
\centering
\caption{Benchmark comparison for the Llama-3.3-70B-Instruct compression pipeline. GSM8K uses strict exact-match; IFEval~\cite{zhou2023ifeval} uses instruction-level strict accuracy; BoolQ~\cite{clark2019boolq} and the remaining accuracies use standard accuracy. Pruned layers: 40--43, 54--57, 60--75. Both the QLoRA recovery and the final SFT were performed for 22k iterations (1 epoch) with batch size 128.}
\label{tab:llama70b-benchmarks}
\small
\setlength{\tabcolsep}{4pt}
\begin{tabular}{lccccccccc}
\toprule
Model & Params & MMLU & HellaSwag & WinoGrande & BoolQ & GSM8K & IFEval & MedMCQA & PPL \\
\midrule
Original & 70.55B & 81.99 & 65.75 & 78.37 & 88.69 & 77.94 & 71.10 & 72.22 & 3.96 \\
Block-pruned & 50.02B & 79.56 & 48.79 & 73.09 & 37.74 & 2.96 & 44.84 & 69.64 & 195.25 \\
Pruned+QLoRA & 50.02B & 79.27 & 59.91 & 72.06 & 88.62 & 65.50 & 69.90 & 67.68 & 9.80 \\
Pruned+QLoRA+FG & 35.28B & 63.10 & 46.29 & 66.38 & 86.36 & 2.81 & 48.44 & 40.86 & 24.36 \\
Pruned+QLoRA+FG+SFT & 35.28B & 73.51 & 55.41 & 73.64 & 89.88 & 80.59  & 81.53 & 59.02 & 12.99 \\
\bottomrule
\end{tabular}
\end{table*}

\section{Conclusion}

We introduced \textbf{FermiGrad}, a global rank-allocation method for SVD-compressed language models. It replaces discrete rank selection with differentiable Fermi gates and jointly optimizes the ranks of all compressed layers under a global parameter constraint using global knowledge distillation loss. Because the calibrated low-rank factors and teacher outputs are computed once and cached, rank optimization trains only a small set of rank parameters. In particular, FermiGrad does not require repeatedly decomposing activations or backpropagating through an SVD. This makes for a cheaper optimization problem and decouples rank allocation from the underlying factorization method, allowing FermiGrad to operate on any calibrated full-rank SVD input. FermiGrad consistently improves over uniform rank allocation and STRS. We further demonstrated the scalability of this approach by incorporating FermiGrad into a compression pipeline for Llama-3.3-70B-Instruct.

We also introduced the \textbf{Targeted Compressor}, which replaces the general distillation objective with a task-specific classification objective. This enables the compressed model to retain substantially more knowledge in a selected domain. In particular, SciQ-targeted compression preserves markedly higher MMLU accuracy than general-purpose allocation, and exceeds the MMLU results reported e.g. by (quantized) Dobi-SVD. This gain represents deliberate specialization: performance on unrelated reasoning tasks is allowed to decrease as capacity is redirected toward the target domain. The Targeted Compressor therefore provides a practical mechanism for deriving compact domain experts when an appropriate task-specific dataset is available.

Finally, we proposed \textbf{PivGa}, a lossless secondary representation of low-rank factors based on pivoted gauge fixing. PivGa removes the parameter overhead introduced by factorization while retaining faster inference than the existing PiFa method. It can be applied after rank selection and is therefore an optional component of FermiGrad-SVD and other SVD-based compression methods.

Taken together, these contributions make SVD-based compression more adaptive, modular, and scalable. 

We note that there is still headroom for improvement. Setting different
compression rates in FermiGrad repeats essentially the same computation
during the first iterations, so one could generate a single
$\vec{\mu}$ trajectory and project the closest points onto the parameter
constraints; the analytical projection formulas are given in
Appendix~\ref{app:proj}. Further work includes combining FermiGrad with improved calibration methods
such as IO-SVD, and characterizing the trade-off between specialization
and general capability more systematically.
A limitation shared by all calibration-based compression methods is their dependence on the choice of data. Principled dataset selection therefore remains an open problem, and it is especially pressing for the Targeted Compressor.

\appendix

\section{\label{app:pivga}Derivation of the PivGa factorization}

Here we give the manipulations omitted in Sec.~\ref{sec:pivga}, leading from the
pivoted QR decomposition to the factorization
$W = C \begin{bmatrix}I_{r} & D\end{bmatrix}\Pi^{-1}$.

After pivoting, the weight matrix is split into skeleton columns and the
remaining columns,
$
W\Pi = \begin{bmatrix}C & W_{\text{rest}}\end{bmatrix},
\qquad C\in\mathbb{R}^{m\times r},\quad W_{\text{rest}}\in\mathbb{R}^{m\times(n-r)},
$
where the $r$ skeleton columns are linearly independent by construction. The QR
decomposition gives
\begin{equation}
W\Pi = \begin{bmatrix}Q_1 & Q_2\end{bmatrix}
\begin{bmatrix}R_{11} & R_{12} \\ 0 & R_{22}\end{bmatrix},
\qquad Q_1^{\top}Q_1 = I_r,
\end{equation}
with $R_{11}\in\mathbb{R}^{r\times r}$ upper triangular.

For a matrix of exact rank $r$ we have $R_{22}=0$, and expanding the block
product yields
\begin{equation}
W\Pi = \begin{bmatrix}Q_1 R_{11} & Q_1 R_{12}\end{bmatrix},
\end{equation}
so that $C = Q_1 R_{11}$ and $W_{\text{rest}} = Q_1 R_{12}$.

We now seek the coefficients $D\in\mathbb{R}^{r\times(n-r)}$ that express the
remaining columns in the skeleton-column basis, $W_{\text{rest}} = C D$.
Substituting the identifications above,
\begin{equation}
C D = W_{\text{rest}}
\;\Longrightarrow\;
Q_1 R_{11} D = Q_1 R_{12}.
\end{equation}
Left-multiplying by $Q_1^{\top}$ and using $Q_1^{\top}Q_1 = I_r$ leaves the
triangular system
\begin{equation}
R_{11} D = R_{12}.
\end{equation}
Since $C$ has full column rank $r$ and $C = Q_1 R_{11}$, the block $R_{11}$ is invertible:
\begin{equation}
D = R_{11}^{-1} R_{12}.
\end{equation}
Because $R_{11}$ is triangular, $D$ is obtained by a triangular solve, avoiding explicit matrix inversion in  
the naive gauge fixing described in Sec.~\ref{sec:pivga}. Collecting the blocks,
\begin{equation}
\begin{aligned}
W\Pi &= \begin{bmatrix}C & C D\end{bmatrix}
      = C\begin{bmatrix}I_r & D\end{bmatrix},\\
\text{i.e.}\quad W &= C\begin{bmatrix}I_r & D\end{bmatrix}\Pi^{-1}.
\end{aligned}
\end{equation}

In practice the full coefficient matrix $\begin{bmatrix}I_r & D\end{bmatrix}$ can
be recovered in a single triangular solve of the augmented system
$R_{11}\begin{bmatrix}I_r & D\end{bmatrix} = \begin{bmatrix}R_{11} & R_{12}\end{bmatrix}$,
whose first block returns $I_r$ (confirming the pivoting) and whose second block
returns $D$. Only $C$, $D$ and the permutation $\Pi$ are stored; the identity
block is implicit, giving the parameter count $r(m+n)-r^2$. Finally, acting on
an input $x$, the factorization reads
$Wx = C\big(x_1 + D\,x_2\big)$, where $\begin{bmatrix}x_1 \\ x_2\end{bmatrix} = \Pi^{-1}x$
with $x_1\in\mathbb{R}^{r}$ and $x_2\in\mathbb{R}^{n-r}$, which is the
forward pass given in Sec.~\ref{sec:pivga}.

\section{\label{app:proj}Projections on constraints}

\subsection{Projection onto a hyperplane}

We seek the point $\bm{y}$ on the hyperplane $\inner{\bm{a}}{\bm{y}}=b$ closest to $\bm{x}$:
\begin{equation}
\min_{\bm{y}} \; \tfrac{1}{2}\norm{\bm{x}-\bm{y}}^2
\quad\text{s.t.}\quad \inner{\bm{a}}{\bm{y}}=b .
\end{equation}
We introduce a Lagrange multiplier $\lambda$,
\begin{equation}
\begin{aligned}
\mathcal{L}(\bm{y},\lambda) &= \tfrac12\norm{\bm{x}-\bm{y}}^2 - \lambda\bigl(\inner{\bm{a}}{\bm{y}}-b\bigr),\\
\nabla_{\bm{y}}\mathcal{L} &= (\bm{y}-\bm{x}) - \lambda\bm{a}=\bm{0},
\end{aligned}
\end{equation}
so that $\bm{y}=\bm{x}+\lambda\bm{a}$. Inserting this into the constraint gives
\begin{equation}
\inner{\bm{a}}{\bm{x}+\lambda\bm{a}}=b
\;\Longrightarrow\;
\lambda=\frac{b-\inner{\bm{a}}{\bm{x}}}{\norm{\bm{a}}^2}.
\end{equation}

\subsection{Projection on the quadratic parameter constraint}

The optimization problem now becomes:
\begin{equation}
\min_{\bm{y}} \; \tfrac{1}{2}\norm{\bm{x}-\bm{y}}^2
\quad\text{s.t.}\quad \inner{\bm{a}}{\bm{y}}+b\,\inner{\bm{y}}{\bm{y}}=c .
\end{equation}
The Lagrangian and its stationarity condition are
\begin{equation}
\begin{aligned}
\mathcal{L} &= \tfrac12\norm{\bm{x}-\bm{y}}^2
-\lambda\bigl(\inner{\bm{a}}{\bm{y}}+b\inner{\bm{y}}{\bm{y}}-c\bigr),\\
\bm{0} &= (\bm{y}-\bm{x})-\lambda\bm{a}-2\lambda b\,\bm{y}.
\end{aligned}
\end{equation}
Solving for $\bm{y}$,
\begin{equation}
\bm{y}\,(1-2b\lambda)=\bm{x}+\lambda\bm{a}
\;\Longrightarrow\;
\;\bm{y}=\frac{\bm{x}+\lambda\bm{a}}{1-2b\lambda}\;
\label{eq:parabola-y}
\end{equation}
Again, substituting this into the constraint after some manipulations gives a quadratic equation:
\begin{equation}
A\lambda^2+B\lambda+C=0.
\end{equation}
with
\begin{equation}
\begin{aligned}
A &= b\,\norm{\bm{a}}^2+4b^2c,\\
B &= -(\norm{\bm{a}}^2+4bc),\\
C &= c-\inner{\bm{a}}{\bm{x}}-b\inner{\bm{x}}{\bm{x}}.
\end{aligned}
\end{equation}
Solving this quadratic equation and picking the root that is closest to $\bm{x}$ gives the solution.

\subsection{Projection onto the full constraint set}

The full constraint set is the intersection $\mathcal{C}=\mathcal{B}\cap\mathcal{P}$ of the box $\mathcal{B}$ and the parameter constraint $\mathcal{P}$. Naive alternating projections (POCS),
\begin{equation}
\bm{x}^{(k+1)} = \proj{\mathcal{B}}\!\bigl(\proj{\mathcal{P}}(\bm{x}^{(k)})\bigr),
\end{equation}
need not converge to the closest point; a better solution is obtained using Dykstra's algorithm~\cite{boyle1986dykstra}.

\section{\label{app:full}Full calibration-dataset results}

Here we report the complete comparison of uniform slicing (Unif), STRS, and FermiGrad (FG) across all four calibration datasets (OpenHermes-2.5, Tulu, GeneralThought-420K, and the MMLU train split). Tab.~\ref{tab:fermigrad_vs_uniform} shows the Tulu and MMLU-train subset.

\begin{table*}[tb]
\centering
\small
\setlength{\tabcolsep}{3pt}
\caption{Benchmark accuracy after compression with uniform SVD-LLM slicing (Unif), sensitivity-based rank allocation (STRS~\cite{Yuan2023_ASVD}), and FermiGrad (FG) for the Llama-3.1-8B-Instruct model, across all four calibration datasets and target densities (fraction of parameters retained). PPL is the word-level perplexity on WikiText (lower is better). For each density and benchmark, the single best result across all three methods and all datasets is shown in bold. GSM8K uses strict-match exact-match; the remaining accuracies use standard accuracy.}
\label{tab:fermigrad_vs_uniform_full}
\begin{tabular}{llccccccccccccccc}
\toprule
\multirow{2}{*}{density} & \multirow{2}{*}{Dataset} & \multicolumn{3}{c}{MMLU} & \multicolumn{3}{c}{HellaSwag} & \multicolumn{3}{c}{WinoGrande} & \multicolumn{3}{c}{GSM8K} & \multicolumn{3}{c}{WikiText-2 PPL} \\
\cmidrule(lr){3-5}\cmidrule(lr){6-8}\cmidrule(lr){9-11}\cmidrule(lr){12-14}\cmidrule(lr){15-17}
 &  & Unif & STRS & FG & Unif & STRS & FG & Unif & STRS & FG & Unif & STRS & FG & Unif & STRS & FG \\
\midrule
$1.0$ & Original & \multicolumn{3}{c}{67.8} & \multicolumn{3}{c}{59.1} & \multicolumn{3}{c}{73.6} & \multicolumn{3}{c}{75.9} & \multicolumn{3}{c}{8.7} \\
\midrule
\multirow{4}{*}{0.9} & OH & 54.8 & 56.2 & 59.3 & 47.3 & 44.7 & 50.6 & 68.7 & 68.3 & 70.2 & 25.1 & 0.0 & 41.3 & 21.7 & 36.9 & \textbf{15.5} \\
 & Tulu & 53.0 & 55.6 & 59.1 & 47.3 & 44.0 & 50.2 & 68.3 & 66.9 & 70.2 & 30.9 & 23.8 & \textbf{46.3} & 23.4 & 41.7 & 16.5 \\
 & MMLU-train & 54.3 & 55.5 & 60.9 & 48.7 & 46.6 & \textbf{52.3} & 71.1 & 70.8 & \textbf{73.2} & 3.6 & 0.0 & 18.3 & 23.0 & 40.2 & \textbf{15.5} \\
 & GT & 55.9 & 57.5 & \textbf{61.7} & 48.1 & 45.4 & 51.8 & 69.7 & 69.9 & 71.6 & 19.7 & 0.0 & 33.7 & 22.6 & 42.8 & 16.5 \\
\midrule
\multirow{4}{*}{0.8} & OH & 44.4 & 45.6 & 52.1 & 38.3 & 36.7 & 43.5 & 62.4 & 63.1 & 68.1 & 5.9 & 0.0 & 13.1 & 52.3 & 114.2 & \textbf{25.7} \\
 & Tulu & 42.3 & 45.6 & 51.9 & 37.8 & 36.1 & 43.7 & 62.2 & 64.2 & 65.9 & 11.3 & 2.9 & \textbf{18.3} & 61.1 & 134.0 & 27.4 \\
 & MMLU-train & 45.4 & 48.7 & 54.2 & 41.1 & 38.3 & \textbf{46.2} & 66.3 & 67.4 & \textbf{70.2} & 0.0 & 0.0 & 0.3 & 56.5 & 143.0 & 26.9 \\
 & GT & 47.2 & 48.9 & \textbf{56.0} & 40.3 & 38.2 & 44.8 & 62.8 & 66.7 & 66.9 & 0.0 & 0.0 & 7.7 & 62.0 & 135.5 & 27.9 \\
\midrule
\multirow{4}{*}{0.7} & OH & 33.0 & 37.1 & 42.4 & 31.0 & 30.8 & 36.3 & 56.3 & 59.0 & 61.3 & 0.8 & 0.0 & 3.9 & 203.7 & 355.4 & \textbf{49.7} \\
 & Tulu & 31.7 & 33.3 & 38.1 & 30.7 & 29.9 & 36.4 & 54.7 & 57.3 & 61.1 & 1.7 & 0.0 & \textbf{5.2} & 265.8 & 448.9 & \textbf{49.7} \\
 & MMLU-train & 36.0 & 39.3 & 42.3 & 34.0 & 31.5 & \textbf{39.4} & 61.3 & 62.1 & \textbf{64.6} & 0.0 & 0.0 & 0.0 & 224.7 & 516.2 & 51.5 \\
 & GT & 34.5 & 31.9 & \textbf{45.1} & 31.9 & 30.5 & 37.0 & 55.8 & 60.0 & 61.8 & 0.0 & 0.0 & 0.2 & 240.7 & 515.6 & 56.3 \\
\midrule
\multirow{4}{*}{0.6} & OH & 26.6 & 27.7 & 27.7 & 27.5 & 27.3 & 30.4 & 51.2 & 53.0 & 54.1 & 0.0 & 0.0 & 0.2 & 820.6 & 1038.7 & 114.8 \\
 & Tulu & 27.1 & 26.5 & 27.8 & 27.3 & 26.8 & 30.7 & 51.9 & 51.4 & 55.1 & 0.5 & 0.0 & \textbf{1.3} & 1119.1 & 1483.1 & \textbf{107.4} \\
 & MMLU-train & 29.1 & 29.5 & \textbf{34.1} & 28.8 & 27.6 & \textbf{32.8} & 54.5 & 56.0 & \textbf{58.6} & 0.0 & 0.0 & 0.0 & 788.1 & 1221.1 & 149.1 \\
 & GT & 27.8 & 26.0 & 33.0 & 27.6 & 27.0 & 31.6 & 52.6 & 53.5 & 54.6 & 0.0 & 0.0 & 0.0 & 1590.6 & 1404.8 & 129.4 \\
\midrule
\multirow{4}{*}{0.5} & OH & 26.0 & 23.5 & 25.4 & 26.6 & 26.4 & 27.9 & 48.2 & 50.8 & 51.1 & 0.0 & 0.0 & 0.0 & 3042.2 & 2578.9 & \textbf{292.4} \\
 & Tulu & 25.8 & 23.0 & 24.6 & 26.6 & 26.2 & 28.3 & 49.5 & 51.9 & 51.2 & 0.0 & 0.0 & 0.0 & 3728.8 & 3206.6 & 337.7 \\
 & MMLU-train & 25.3 & 24.8 & \textbf{28.4} & 26.5 & 26.2 & \textbf{28.8} & 52.2 & \textbf{53.4} & 52.5 & 0.0 & 0.0 & 0.0 & 2815.9 & 2332.4 & 504.5 \\
 & GT & 24.5 & 24.1 & 24.2 & 26.4 & 26.4 & 27.6 & 50.8 & 51.5 & 52.0 & 0.0 & 0.0 & 0.0 & 4504.0 & 3372.7 & 554.7 \\
\bottomrule
\end{tabular}
\end{table*}

\section{\label{app:balance}Gauge balancing of the low-rank factors}

After compression each weight matrix is stored as a low-rank product $W=uv$ with $u\in\mathbb{R}^{m\times r}$ and $v\in\mathbb{R}^{r\times n}$ (the factors $A,B$ of Sec.~\ref{sec:svd}),
\begin{equation}
W_{ik}=\sum_{j=0}^{r-1} u_{ij}\,v_{jk}.
\end{equation}
Because of the gauge freedom between the two factors, it is not guaranteed that $u$ and $v$ are on the same scale (e.g. it is in principle possible to scale $u$ by $s\gg1$ and $v$ by $1/s$ without altering the forward pass). An imbalance is detrimental during retraining, where both factors share a single learning rate and optimizer state: the large-scale factor dominates the updates while the small-scale factor barely moves, which makes the training unstable. We therefore rebalance the two factors before retraining.

Limiting ourselves to a diagonal gauge,
\begin{equation}
W_{ik}=\sum_{j=0}^{r-1} u_{ij}\, d_j^{-1} d_j\, v_{jk},
\end{equation}
which amounts to rescaling the $j$-th column of $u$ by $d_j^{-1}$ and the $j$-th row of $v$ by $d_j$, we use it to balance the scales of the two factors.

As the scale measure we take the root-mean-square (RMS) of each rank component, i.e.\ of each column of $u$ and each row of $v$,
\begin{equation}
\operatorname{rms}^2_{\mathrm{col},j}(u)
= \frac{1}{m} \sum_{i=0}^{m-1} u_{ij}^2 = \frac{1}{m} \lr{u^T u}_{jj},
\end{equation}
\begin{equation}
\operatorname{rms}^2_{\mathrm{row},j}(v)
= \frac{1}{n} \sum_{k=0}^{n-1} v_{jk}^2 = \frac{1}{n} \lr{v v^T}_{jj}.
\end{equation}
Choosing the gauge
\begin{equation}
d_j = \sqrt{\frac{\operatorname{rms}_{\mathrm{col},j}(u)}
                 {\operatorname{rms}_{\mathrm{row},j}(v)}}
\end{equation}
and forming the balanced factors
\begin{align}
u^{\mathrm{bal}}_{ij} &= d_j^{-1}\, u_{ij}, &
v^{\mathrm{bal}}_{jk} &= d_j\, v_{jk},
\end{align}
leaves the product unchanged, $u^{\mathrm{bal}} v^{\mathrm{bal}} = u v = W$, while equalizing the two scales for every component,
\begin{equation}
\begin{split}
\operatorname{rms}_{\mathrm{col},j}\!\big(u^{\mathrm{bal}}\big)
&= \operatorname{rms}_{\mathrm{row},j}\!\big(v^{\mathrm{bal}}\big) \\
&= \sqrt{ \operatorname{rms}_{\mathrm{col},j}(u) \,
         \operatorname{rms}_{\mathrm{row},j}(v) },
\end{split}
\end{equation}
the geometric mean of the original scales.

Although the balancing is derived independently for each rank component $j$, we empirically find that it also equalizes the \emph{global} weight distributions of the two factors: after balancing, $u$ and $v$ have essentially the same overall RMS.

\bibliographystyle{plainnat}
\bibliography{FermiGrad}

@inproceedings{Zhao2025_Pifa,
  title         = {Pivoting Factorization: A Compact Meta Low-Rank Representation of Sparsity for Efficient Inference in Large Language Models},
  author        = {Zhao, Jialin and Zhang, Yingtao and Cannistraci, Carlo Vittorio},
  booktitle     = {Proceedings of the 42nd International Conference on Machine Learning},
  series        = {Proceedings of Machine Learning Research},
  volume        = {267},
  pages         = {77926--77947},
  publisher     = {PMLR},
  year          = {2025},
  eprint        = {2501.19090},
  archivePrefix = {arXiv},
  primaryClass  = {cs.LG},
  url           = {https://proceedings.mlr.press/v267/zhao25af.html}
}

@inproceedings{ParkinaRakhuba2025_COALA,
  title         = {{COALA}: Numerically Stable and Efficient Framework for Context-Aware Low-Rank Approximation},
  author        = {Parkina, Uliana and Rakhuba, Maxim},
  booktitle     = {Advances in Neural Information Processing Systems},
  volume        = {38},
  year          = {2025},
  eprint        = {2507.07580},
  archivePrefix = {arXiv},
  primaryClass  = {cs.LG},
  url           = {https://proceedings.neurips.cc/paper_files/paper/2025/hash/67022e6927b7887ef56f57d87d42ca84-Abstract-Conference.html}
}

@inproceedings{Koike-Akino2025_LatentLLM,
  title         = {Latent{LLM}: Attention-Aware Joint Tensor Compression},
  author        = {Koike-Akino, Toshiaki and Chen, Xiangyu and Liu, Jing and Wang, Ye and Wang, Pu and Brand, Matthew},
  booktitle     = {IEEE Conference on Computer Vision and Pattern Recognition (CVPR) Workshop},
  year          = {2025},
  month         = jun,
  eprint        = {2505.18413},
  archivePrefix = {arXiv},
  primaryClass  = {cs.LG},
  url           = {https://www.merl.com/publications/TR2025-075}
}

@inproceedings{Gromov2024_Unreasonable,
  title         = {The Unreasonable Ineffectiveness of the Deeper Layers},
  author        = {Gromov, Andrey and Tirumala, Kushal and Shapourian, Hassan and Glorioso, Paolo and Roberts, Daniel A.},
  booktitle     = {The Thirteenth International Conference on Learning Representations},
  year          = {2025},
  eprint        = {2403.17887},
  archivePrefix = {arXiv},
  primaryClass  = {cs.CL},
  url           = {https://proceedings.iclr.cc/paper_files/paper/2025/hash/cbabc2f70de2dd09f491a8715ec3e80f-Abstract-Conference.html}
}

@misc{Sreenivas2024_Minitron,
  title         = {{LLM} Pruning and Distillation in Practice: The Minitron Approach},
  author        = {Sreenivas, Sharath Turuvekere and Muralidharan, Saurav and Joshi, Raviraj and Chochowski, Marcin and Mahabaleshwarkar, Ameya Sunil and Shen, Gerald and Zeng, Jiaqi and Chen, Zijia and Suhara, Yoshi and Diao, Shizhe and Yu, Chenhan and Chen, Wei-Chun and Ross, Hayley and Olabiyi, Oluwatobi and Aithal, Ashwath and Kuchaiev, Oleksii and Korzekwa, Daniel and Molchanov, Pavlo and Patwary, Mostofa and Shoeybi, Mohammad and Kautz, Jan and Catanzaro, Bryan},
  year          = {2024},
  eprint        = {2408.11796},
  archivePrefix = {arXiv},
  primaryClass  = {cs.CL},
  doi           = {10.48550/arXiv.2408.11796},
  url           = {https://arxiv.org/abs/2408.11796}
}

@inproceedings{Dettmers2022_quant,
  title         = {{LLM}.int8(): 8-bit Matrix Multiplication for Transformers at Scale},
  author        = {Dettmers, Tim and Lewis, Mike and Belkada, Younes and Zettlemoyer, Luke},
  booktitle     = {Advances in Neural Information Processing Systems},
  volume        = {35},
  year          = {2022},
  eprint        = {2208.07339},
  archivePrefix = {arXiv},
  primaryClass  = {cs.CL},
  url           = {https://papers.neurips.cc/paper_files/paper/2022/hash/c3ba4962c05c49636d4c6206a97e9c8a-Abstract-Conference.html}
}

@inproceedings{Wang2024_SVD-LLM,
  title         = {{SVD-LLM}: Truncation-aware Singular Value Decomposition for Large Language Model Compression},
  author        = {Wang, Xin and Zheng, Yu and Wan, Zhongwei and Zhang, Mi},
  booktitle     = {The Thirteenth International Conference on Learning Representations},
  year          = {2025},
  eprint        = {2403.07378},
  archivePrefix = {arXiv},
  primaryClass  = {cs.CL},
  url           = {https://proceedings.iclr.cc/paper_files/paper/2025/hash/3104e1ab39875cf54fe1eb4473e7c5a1-Abstract-Conference.html}
}

@misc{Yuan2023_ASVD,
  title         = {{ASVD}: Activation-aware Singular Value Decomposition for Compressing Large Language Models},
  author        = {Yuan, Zhihang and Shang, Yuzhang and Song, Yue and Yang, Dawei and Wu, Qiang and Yan, Yan and Sun, Guangyu},
  year          = {2023},
  eprint        = {2312.05821},
  archivePrefix = {arXiv},
  primaryClass  = {cs.CL},
  doi           = {10.48550/arXiv.2312.05821},
  url           = {https://arxiv.org/abs/2312.05821}
}

@misc{jansen2026,
  title         = {{LLM} Compression by Block Removal with Constrained Binary Optimization},
  author        = {Jansen, David and Rausch, Roman and Hashemi, Ali and Montero, David and Or{\'u}s, Rom{\'a}n},
  year          = {2026},
  eprint        = {2602.00161},
  archivePrefix = {arXiv},
  primaryClass  = {cs.LG},
  doi           = {10.48550/arXiv.2602.00161},
  url           = {https://arxiv.org/abs/2602.00161}
}

@inproceedings{lin2023awq,
  title         = {{AWQ}: Activation-aware Weight Quantization for On-Device {LLM} Compression and Acceleration},
  author        = {Lin, Ji and Tang, Jiaming and Tang, Haotian and Yang, Shang and Chen, Wei-Ming and Wang, Wei-Chen and Xiao, Guangxuan and Dang, Xingyu and Gan, Chuang and Han, Song},
  booktitle     = {Proceedings of Machine Learning and Systems},
  volume        = {6},
  pages         = {87--100},
  year          = {2024},
  eprint        = {2306.00978},
  archivePrefix = {arXiv},
  primaryClass  = {cs.CL},
  url           = {https://proceedings.mlsys.org/paper_files/paper/2024/hash/42a452cbafa9dd64e9ba4aa95cc1ef21-Abstract-Conference.html}
}

@inproceedings{frantar2022gptq,
  title         = {{OPTQ}: Accurate Post-Training Quantization for Generative Pre-trained Transformers},
  author        = {Frantar, Elias and Ashkboos, Saleh and Hoefler, Torsten and Alistarh, Dan},
  booktitle     = {The Eleventh International Conference on Learning Representations},
  year          = {2023},
  eprint        = {2210.17323},
  archivePrefix = {arXiv},
  primaryClass  = {cs.LG},
  url           = {https://openreview.net/forum?id=tcbBPnfwxS},
  note          = {The arXiv version is titled GPTQ: Accurate Post-Training Quantization for Generative Pre-trained Transformers.}
}

@inproceedings{sun2023wanda,
  title         = {A Simple and Effective Pruning Approach for Large Language Models},
  author        = {Sun, Mingjie and Liu, Zhuang and Bair, Anna and Kolter, J. Zico},
  booktitle     = {The Twelfth International Conference on Learning Representations},
  year          = {2024},
  eprint        = {2306.11695},
  archivePrefix = {arXiv},
  primaryClass  = {cs.LG},
  url           = {https://proceedings.iclr.cc/paper_files/paper/2024/hash/14c856c7a41297804de4c4890e846b25-Abstract-Conference.html}
}

@inproceedings{muralidharan_24,
  title         = {Compact Language Models via Pruning and Knowledge Distillation},
  author        = {Muralidharan, Saurav and Sreenivas, Sharath Turuvekere and Joshi, Raviraj and Chochowski, Marcin and Patwary, Mostofa and Shoeybi, Mohammad and Catanzaro, Bryan and Kautz, Jan and Molchanov, Pavlo},
  booktitle     = {Advances in Neural Information Processing Systems},
  volume        = {37},
  year          = {2024},
  doi           = {10.52202/079017-1299},
  eprint        = {2407.14679},
  archivePrefix = {arXiv},
  primaryClass  = {cs.CL},
  url           = {https://papers.nips.cc/paper_files/paper/2024/hash/4822991365c962105b1b95b1107d30e5-Abstract-Conference.html}
}

@inproceedings{guo2025slimllm,
  title         = {Slim{LLM}: Accurate Structured Pruning for Large Language Models},
  author        = {Guo, Jialong and Chen, Xinghao and Tang, Yehui and Wang, Yunhe},
  booktitle     = {Proceedings of the 42nd International Conference on Machine Learning},
  series        = {Proceedings of Machine Learning Research},
  volume        = {267},
  pages         = {20766--20776},
  publisher     = {PMLR},
  year          = {2025},
  eprint        = {2505.22689},
  archivePrefix = {arXiv},
  primaryClass  = {cs.LG},
  url           = {https://proceedings.mlr.press/v267/guo25a.html}
}

@misc{Gili_25,
  title         = {Scalable Iterative Pruning of Large Language and Vision Models Using Block Coordinate Descent},
  author        = {Rosenberg, Gili and Brubaker, J. Kyle and Schuetz, Martin J. A. and Zhu, Elton Yechao and Kad{\i}o{\u{g}}lu, Serdar and Borujeni, Sima E. and Katzgraber, Helmut G.},
  year          = {2024},
  eprint        = {2411.17796},
  archivePrefix = {arXiv},
  primaryClass  = {cs.LG},
  doi           = {10.48550/arXiv.2411.17796},
  url           = {https://arxiv.org/abs/2411.17796}
}

@inproceedings{Wang2025_DobiSVD,
  title         = {Dobi-{SVD}: Differentiable {SVD} for {LLM} Compression and Some New Perspectives},
  author        = {Wang, Qinsi and Ke, Jinghan and Tomizuka, Masayoshi and Keutzer, Kurt and Xu, Chenfeng},
  booktitle     = {The Thirteenth International Conference on Learning Representations},
  year          = {2025},
  eprint        = {2502.02723},
  archivePrefix = {arXiv},
  primaryClass  = {cs.LG},
  url           = {https://proceedings.iclr.cc/paper_files/paper/2025/hash/218ca0d92e6ed8f9db00621e103dc70c-Abstract-Conference.html}
}

@inproceedings{welbl2017sciq,
  title         = {Crowdsourcing Multiple Choice Science Questions},
  author        = {Welbl, Johannes and Liu, Nelson F. and Gardner, Matt},
  booktitle     = {Proceedings of the 3rd Workshop on Noisy User-generated Text},
  pages         = {94--106},
  year          = {2017},
  publisher     = {Association for Computational Linguistics},
  address       = {Copenhagen, Denmark},
  doi           = {10.18653/v1/W17-4413},
  url           = {https://aclanthology.org/W17-4413/}
}

@inproceedings{pal2022medmcqa,
  title         = {{MedMCQA}: A Large-scale Multi-Subject Multi-Choice Dataset for Medical Domain Question Answering},
  author        = {Pal, Ankit and Umapathi, Logesh Kumar and Sankarasubbu, Malaikannan},
  booktitle     = {Proceedings of the Conference on Health, Inference, and Learning},
  series        = {Proceedings of Machine Learning Research},
  volume        = {174},
  pages         = {248--260},
  year          = {2022},
  publisher     = {PMLR},
  eprint        = {2203.14371},
  archivePrefix = {arXiv},
  primaryClass  = {cs.CL},
  url           = {https://proceedings.mlr.press/v174/pal22a.html}
}

@misc{teknium2023openhermes,
  title         = {OpenHermes 2.5: An Open Dataset of Synthetic Data for Generalist {LLM} Assistants},
  author        = {{Teknium}},
  year          = {2023},
  publisher     = {Hugging Face},
  howpublished  = {Hugging Face dataset},
  url           = {https://huggingface.co/datasets/teknium/OpenHermes-2.5}
}

@inproceedings{hendrycks2021mmlu,
  title         = {Measuring Massive Multitask Language Understanding},
  author        = {Hendrycks, Dan and Burns, Collin and Basart, Steven and Zou, Andy and Mazeika, Mantas and Song, Dawn and Steinhardt, Jacob},
  booktitle     = {International Conference on Learning Representations},
  year          = {2021},
  eprint        = {2009.03300},
  archivePrefix = {arXiv},
  primaryClass  = {cs.CY},
  url           = {https://openreview.net/forum?id=d7KBjmI3GmQ},
  note          = {Dataset: https://huggingface.co/datasets/cais/mmlu}
}

@inproceedings{muennighoff2024olmoe,
  title         = {{OLMoE}: Open Mixture-of-Experts Language Models},
  author        = {Muennighoff, Niklas and Soldaini, Luca and Groeneveld, Dirk and Lo, Kyle and Morrison, Jacob and Min, Sewon and Shi, Weijia and Walsh, Pete and Tafjord, Oyvind and Lambert, Nathan and Gu, Yuling and Arora, Shane and Bhagia, Akshita and Schwenk, Dustin and Wadden, David and Wettig, Alexander and Hui, Binyuan and Dettmers, Tim and Kiela, Douwe and Farhadi, Ali and Smith, Noah A. and Koh, Pang Wei and Singh, Amanpreet and Hajishirzi, Hanna},
  booktitle     = {The Thirteenth International Conference on Learning Representations},
  year          = {2025},
  eprint        = {2409.02060},
  archivePrefix = {arXiv},
  primaryClass  = {cs.CL},
  url           = {https://proceedings.iclr.cc/paper_files/paper/2025/hash/9b224ace8963c9385ad5e2b5c9039b97-Abstract-Conference.html},
  note          = {SFT mix: https://huggingface.co/datasets/allenai/tulu-v3.1-mix-preview-4096-OLMoE}
}

@inproceedings{FWSVD,
  title         = {Language Model Compression with Weighted Low-Rank Factorization},
  author        = {Hsu, Yen-Chang and Hua, Ting and Chang, Sungen and Lou, Qian and Shen, Yilin and Jin, Hongxia},
  booktitle     = {International Conference on Learning Representations},
  year          = {2022},
  eprint        = {2207.00112},
  archivePrefix = {arXiv},
  primaryClass  = {cs.CL},
  url           = {https://openreview.net/forum?id=uPv9Y3gmAI5}
}

@misc{abbasi2026iosvd,
  title         = {{IO-SVD}: Input-Output Whitened {SVD} for Adaptive-Rank {LLM} Compression},
  author        = {Abbasi, Ali and Thrash, Chayne and Qin, Haoran and Pirsiavash, Hamed and Kolouri, Soheil},
  year          = {2026},
  eprint        = {2605.15626},
  archivePrefix = {arXiv},
  primaryClass  = {cs.LG},
  doi           = {10.48550/arXiv.2605.15626},
  url           = {https://arxiv.org/abs/2605.15626}
}

@inproceedings{Novikov2015_TensorizingNN,
  title         = {Tensorizing Neural Networks},
  author        = {Novikov, Alexander and Podoprikhin, Dmitrii and Osokin, Anton and Vetrov, Dmitry P.},
  booktitle     = {Advances in Neural Information Processing Systems},
  volume        = {28},
  year          = {2015},
  eprint        = {1509.06569},
  archivePrefix = {arXiv},
  primaryClass  = {cs.LG},
  url           = {https://proceedings.neurips.cc/paper/2015/hash/6855456e2fe46a9d49d3d3af4f57443d-Abstract.html}
}

@misc{generalthought2025,
  title         = {{GeneralThought-430K}},
  author        = {{RJT1990}},
  year          = {2025},
  publisher     = {Hugging Face},
  howpublished  = {Hugging Face dataset},
  url           = {https://huggingface.co/datasets/RJT1990/GeneralThoughtArchive},
  note          = {Archived dataset release dated March 14, 2025}
}

@inproceedings{merity2017wikitext,
  title         = {Pointer Sentinel Mixture Models},
  author        = {Merity, Stephen and Xiong, Caiming and Bradbury, James and Socher, Richard},
  booktitle     = {International Conference on Learning Representations},
  year          = {2017},
  eprint        = {1609.07843},
  archivePrefix = {arXiv},
  primaryClass  = {cs.CL},
  url           = {https://openreview.net/forum?id=Byj72udxe}
}

@article{jin2020disease,
  title         = {What Disease Does This Patient Have? A Large-Scale Open Domain Question Answering Dataset from Medical Exams},
  author        = {Jin, Di and Pan, Eileen and Oufattole, Nassim and Weng, Wei-Hung and Fang, Hanyi and Szolovits, Peter},
  journal       = {Applied Sciences},
  volume        = {11},
  number        = {14},
  pages         = {6421},
  year          = {2021},
  publisher     = {MDPI},
  doi           = {10.3390/app11146421},
  eprint        = {2009.13081},
  archivePrefix = {arXiv},
  primaryClass  = {cs.CL},
  url           = {https://www.mdpi.com/2076-3417/11/14/6421}
}

@misc{nemo_automodel,
  title        = {{NeMo} {AutoModel}: {PyTorch}-Native Distributed Training Library with Day-0 {Hugging Face} Support},
  author       = {{NVIDIA}},
  year         = {2025},
  howpublished = {\url{https://github.com/NVIDIA-NeMo/Automodel}},
  note         = {NVIDIA NeMo Framework, software repository}
}

@inproceedings{zellers2019hellaswag,
  title         = {{HellaSwag}: Can a Machine Really Finish Your Sentence?},
  author        = {Zellers, Rowan and Holtzman, Ari and Bisk, Yonatan and Farhadi, Ali and Choi, Yejin},
  booktitle     = {Proceedings of the 57th Annual Meeting of the Association for Computational Linguistics},
  pages         = {4791--4800},
  year          = {2019},
  publisher     = {Association for Computational Linguistics},
  address       = {Florence, Italy},
  doi           = {10.18653/v1/P19-1472},
  eprint        = {1905.07830},
  archivePrefix = {arXiv},
  primaryClass  = {cs.CL},
  url           = {https://aclanthology.org/P19-1472/}
}

@inproceedings{sakaguchi2020winogrande,
  title         = {{WinoGrande}: An Adversarial Winograd Schema Challenge at Scale},
  author        = {Sakaguchi, Keisuke and Le Bras, Ronan and Bhagavatula, Chandra and Choi, Yejin},
  booktitle     = {Proceedings of the AAAI Conference on Artificial Intelligence},
  volume        = {34},
  pages         = {8732--8740},
  year          = {2020},
  doi           = {10.1609/aaai.v34i05.6399},
  eprint        = {1907.10641},
  archivePrefix = {arXiv},
  primaryClass  = {cs.CL},
  url           = {https://ojs.aaai.org/index.php/AAAI/article/view/6399}
}

@misc{cobbe2021gsm8k,
  title         = {Training Verifiers to Solve Math Word Problems},
  author        = {Cobbe, Karl and Kosaraju, Vineet and Bavarian, Mohammad and Chen, Mark and Jun, Heewoo and Kaiser, Lukasz and Plappert, Matthias and Tworek, Jerry and Hilton, Jacob and Nakano, Reiichiro and Hesse, Christopher and Schulman, John},
  year          = {2021},
  eprint        = {2110.14168},
  archivePrefix = {arXiv},
  primaryClass  = {cs.LG},
  doi           = {10.48550/arXiv.2110.14168},
  url           = {https://arxiv.org/abs/2110.14168}
}

@inproceedings{clark2019boolq,
  title         = {{BoolQ}: Exploring the Surprising Difficulty of Natural Yes/No Questions},
  author        = {Clark, Christopher and Lee, Kenton and Chang, Ming-Wei and Kwiatkowski, Tom and Collins, Michael and Toutanova, Kristina},
  booktitle     = {Proceedings of the 2019 Conference of the North American Chapter of the Association for Computational Linguistics: Human Language Technologies, Volume 1 (Long and Short Papers)},
  pages         = {2924--2936},
  year          = {2019},
  publisher     = {Association for Computational Linguistics},
  address       = {Minneapolis, Minnesota},
  doi           = {10.18653/v1/N19-1300},
  url           = {https://aclanthology.org/N19-1300/}
}

@misc{zhou2023ifeval,
  title         = {Instruction-Following Evaluation for Large Language Models},
  author        = {Zhou, Jeffrey and Lu, Tianjian and Mishra, Swaroop and Brahma, Siddhartha and Basu, Sujoy and Luan, Yi and Zhou, Denny and Hou, Le},
  year          = {2023},
  eprint        = {2311.07911},
  archivePrefix = {arXiv},
  primaryClass  = {cs.CL},
  doi           = {10.48550/arXiv.2311.07911},
  url           = {https://arxiv.org/abs/2311.07911}
}

@misc{fan2025megascience,
  title         = {{MegaScience}: Pushing the Frontiers of Post-Training Datasets for Science Reasoning},
  author        = {Fan, Run-Ze and Wang, Zengzhi and Liu, Pengfei},
  year          = {2025},
  eprint        = {2507.16812},
  archivePrefix = {arXiv},
  primaryClass  = {cs.CL},
  doi           = {10.48550/arXiv.2507.16812},
  url           = {https://arxiv.org/abs/2507.16812},
  note          = {Includes the TextbookReasoning dataset}
}

@incollection{boyle1986dykstra,
  title         = {A Method for Finding Projections onto the Intersection of Convex Sets in {H}ilbert Spaces},
  author        = {Boyle, James P. and Dykstra, Richard L.},
  booktitle     = {Advances in Order Restricted Statistical Inference},
  series        = {Lecture Notes in Statistics},
  volume        = {37},
  pages         = {28--47},
  publisher     = {Springer},
  address       = {New York, NY},
  year          = {1986},
  doi           = {10.1007/978-1-4613-9940-7_3},
  url           = {https://link.springer.com/chapter/10.1007/978-1-4613-9940-7_3}
}

@misc{lm-eval2021,
  title        = {A Framework for Few-Shot Language Model Evaluation},
  author       = {Gao, Leo and Tow, Jonathan and Biderman, Stella and Black, Sid and DiPofi, Anthony and Foster, Charles and Golding, Laurence and Hsu, Jeffrey and McDonell, Kyle and Muennighoff, Niklas and Phang, Jason and Reynolds, Laria and Tang, Eric and Thite, Anish and Wang, Ben and Wang, Kevin and Zou, Andy},
  year         = {2021},
  howpublished = {Zenodo, \url{https://doi.org/10.5281/zenodo.5371628}},
  note         = {lm-evaluation-harness, v0.0.1}
}

\end{document}